\documentclass[conference]{IEEEtran}
\IEEEoverridecommandlockouts
\usepackage{cite}
\usepackage{amsmath,amssymb,amsfonts}
\usepackage{algpseudocode}

\usepackage{algorithm}
\usepackage{graphicx}
\usepackage{textcomp}
\usepackage{xcolor}
\usepackage{comment}
\def\BibTeX{{\rm B\kern-.05em{\sc i\kern-.025em b}\kern-.08em
		T\kern-.1667em\lower.7ex\hbox{E}\kern-.125emX}}

\def\BibTeX{{\rm B\kern-.05em{\sc i\kern-.025em b}\kern-.08em
		T\kern-.1667em\lower.7ex\hbox{E}\kern-.125emX}}

\begin{document}
\title{Accelerating Path Planning for Autonomous Driving with Hardware-Assisted 
	Memoization}



\author{\IEEEauthorblockN{Mulong Luo}
	\IEEEauthorblockA{
		\textit{Cornell University}\\
		Ithaca, NY, USA \\
		ml2558@cornell.edu}\footnote{This project is partially funded by NSF grant 
		ECCS-1932501.}
	\and
	\IEEEauthorblockN{G. Edward Suh}
	\IEEEauthorblockA{
	\textit{Cornell University}\\
	Ithaca, NY, USA \\
	suh@ece.cornell.edu}
}
	
\maketitle
\pagestyle{plain}


\begin{abstract}

Path planning for autonomous driving with dynamic obstacles poses a challenge because
it needs to perform a higher-dimensional search (with time-dimension) while still meeting
real-time constraints. 
This paper proposes an algorithm-hardware co-optimization approach to 
accelerate path planning with high-dimensional search space. 
First, we reduce the time for a nearest neighbor search and collision 
detection by mapping nodes and obstacles to a lower-dimensional space and 
memoizing recent search results.
Then, we propose a 
hardware extension for efficient memoization. 
The experimental results on a modern processor and a cycle-level simulator show
that the hardware-assisted memoization significantly reduces the execution time of path planning.

\end{abstract}

\section{Introduction}

 


Path planning for autonomous driving faces challenges including
strict real-time constraints as well as  dynamic obstacles.
First, in autonomous driving, path planning can take 100 $ms$  on average on a 
commercial  autonomous driving development platform \cite{yu2020building} like 
Baidu Apollo, while  the safety 
requirement assumes the 
sensor-to-actuator latency must be less than 100 $ms$ \cite{lin2018architectural} to ensure 
the 
responsiveness.  
Second, modern autonomous vehicles have to deal with 
dynamic obstacles. In real-world streets, there are other vehicles moving at high 
speed, and there are pedestrians walking at low speed. Path planning algorithm has 
to deal 
with these moving objects and avoid collisions. The vehicles
have to take into account the locations of the obstacles at different timesteps.
Recent studies on path planning algorithms 
such as PUMP \cite{ichter2017robust} and contingency 
planning \cite{hardy2013contingency} consider dynamic obstacles. However, it  
could take as long as 500 $ms$ to compute 
a path in these scenarios on a CPU. 
Even on a GPU, PUMP still needs around 100 $ms$ to find 
a path.


On-time execution of path planning is important for the safety and 
the efficiency of a vehicle. 
Pruning \cite{pan2016fast} has been used to accelerate path planning in software.
Algorithm-specific accelerators have also been proposed
for rapid-exploring random tree (RRT) \cite{xiao2017parallel}, probablistic 
roadmap (PRM)\cite{murray2019programmable}, and 
A* \cite{kim2017brain}, using  FPGAs or 
ASICs. While 
these accelerators for path planning do address the 
individual needs in the specific scenarios, they do not target the autonomous 
driving and do not consider dynamic obstacles.

Besides, the flexibility to choose and modify the path planning algorithm to be 
accelerated  is also very important, as the 
regulations and 
algorithms are evolving continuously. However, many of the existing path 
planning accelerator designs only consider one specific algorithm \cite{xiao2017parallel} at a time with 
most of the hyper-parameters baked into the accelerator hardware at design/program time. While these 
hardwired algorithms 
do perform well and provide good performance on the targeted scenario at the time 
of 
launch, they may not be flexible enough to adapt to new or modified algorithms.

In this paper, we address  the problem of path planning for autonomous driving with dynamic 
obstacles.
The major contributions of 
this work is as follows:

\begin{enumerate}
\item We introduce the  space-filling curve for efficient indexing, 
memoization and pruning of the time-consuming nearest neighbor search and collision 
detection in path planning.

\item We propose a high-performance hardware implementation and a 
programming interface for memoization of the space-filling curve-indexed tree nodes, which 
is used for a nearest neighbor search  and collision detection.

\item We demonstrate that the proposed hardware-assisted approach can lead to 
significant performance improvement for path planning with dynamic obstacles in both synthetic and 
realistic benchmarks.
	
\end{enumerate}

\section{Approach}\label{sec:design}
 
 


\noindent{\bf Handling Dynamic Obstacles. }
Let us consider path planning for an autonomous vehicle on a 2D plane. At different times, 
obstacles will be at different positions on the 2D plane.  A dynamic obstacle in a 2D plane becomes a 
set 
of obstacles 
whose coordinates in the 3D space can be described by $(x, y, t)$. We can effectively regard 
the 2D 
planning problem with  dynamic obstacles as a 3D planning problem.
However, the time dimension is different from a normal spatial dimension and introduces 
an additional 
constraint. Since time only goes in one direction, for any two nodes $(x_i, y_i, t_i)$ and 
$(x_j, y_j, t_j)$ on a path P where $i < j$, we must guarantee $t_i < t_j$. The additional constraint is 
enforced when selecting nearest neighbors in RRT. 

\begin{figure}[h]
  		\vspace{-0.5cm}
	\centering
	\includegraphics[width=0.6\columnwidth]{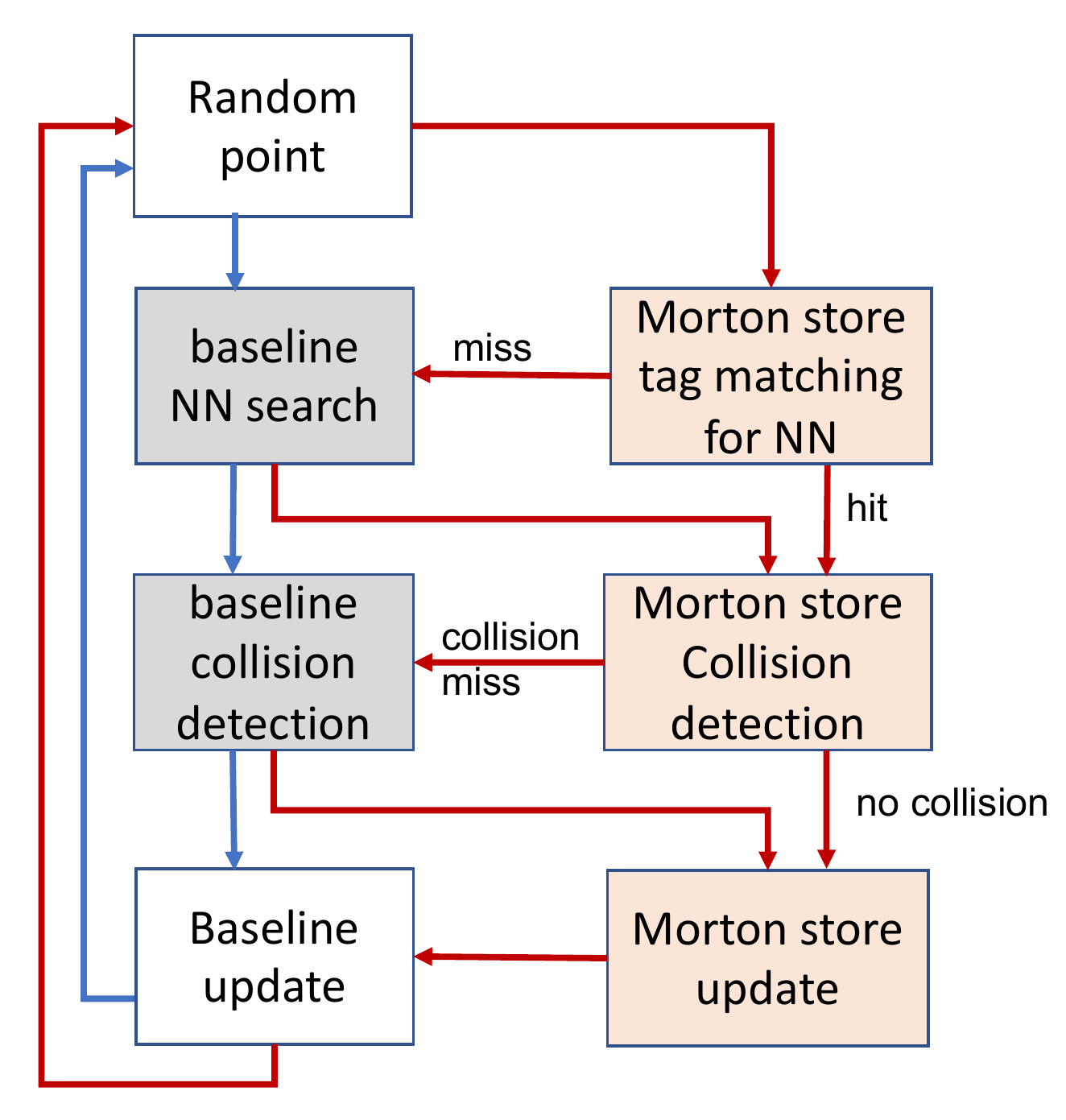}
	\vspace{-0.25cm}
	\caption{The baseline (blue arrows) and proposed (orange arrows) planning 
	flow.}\label{fig:flow}
			\vspace{-0.25cm}
\end{figure}

\noindent{\bf Software-based Memoization. }
We use memoization to reduce the total execution time of path planning. 
In traditional sampling-based planning, 
a large portion of the execution time is spent on finding the nearest neighbor (NN) and detecting collisions.  To 
accelerate these 
 procedures, we memoize the recently accessed nodes and the collision state so that some of 
 the similar queries  can be 
 skipped. 
 Figure~\ref{fig:flow} shows the high-level flow of our approach. In the 
baseline RRT, the time-consuming nearest neighbor search and collision detection are performed 
for all iterations. 
However, we use an small-size data storage named {\em Morton store} (named after Morton space-filing 
curve \cite{open2017discrete}) to 
memoize the key 
information from 
baseline nearest neighbor search and collision detection. For each iteration, we always check the Morton store first 
to 
opportunistically skip the time-consuming baseline NN search and collision detection.

The  Morton store works as follows.
We first calculate the Morton codes $M_n$  and $M_o$ for the coordinates of both tree nodes 
$X_n=(x_n, y_n, t_n)$ 
and 
obstacles $X_o=(x_o, y_o, t_o)$. 
Morton code projects point $(x, y, t)$  from 3D space onto a 1D curve, while maintaining spatial 
locality ($x$, $y$, $t$ are represented as 32-bit integers).
The Morton codes $M_n$  and $M_o$  are 64-bit numbers and are used as the tags for the 
corresponding nodes and 
obstacles. 
To adjust the projection granularity, we mask $k$ least significant bits of the Morton code 
\begin{equation}
	M_n' \gets M_n\; \texttt{\&} \; \underbrace{1...1}_{(64-k)\; bits}\underbrace{0...0}_{k\; bits}
\end{equation}


The pseudocode for the Morton-store based acceleration approach is shown in Algorithm~\ref{alg:cap}.
For each iteration we first generate  a new random node (Line~2). Then we calculate its masked Morton 
code and 
search for entries with matching Morton codes in the Morton store (Line~3). If there is no match in 
the Morton store,  we perform the baseline nearest neighbor search using 
kdtree (Line~4). Then we perform the ``steer'' operation to find a node along the direction of 
$X_{rand}$ but with unit distance to $X_{nearest}$.  For 
collision detection, we also check the Morton store first (Line~8). 
If the result indicates that it is not NO\_COLLISION state (could be COLLISION state or a miss), we do 
more 
time-consuming baseline detailed 
collision detection (Line~9). After that, we update the 
Morton store (Line~11) and the baseline tree (Line~12-14).
The solution path has to perform exact collision detection for each segment to make sure the path 
is indeed safe (NO\_COLLISION). Compared with the time saved by not performing exact collision detection 
using the Morton store, the 
solution checking overhead is negligible. The time complexity of the algorithm is 
$O(N\cdot(\alpha \cdot log N+\beta\cdot  log L))$ where $N$ is the number of nodes in the tree, $L$ is 
the 
number of obstacles, and $\alpha$, $\beta$ are probabilities Line~8 and Line~12 are taken, respectively 
($0< 
\alpha, \beta\leq 1$). For normal RRT without using the Morton store, we have $\alpha, \beta = 
1$.

\begin{algorithm}
	\scriptsize
	\caption{RRT with Morton code-based memoization}\label{alg:cap}
	\begin{algorithmic}[1]
		\While {$X_{goal}$ not reached}
		\State {$X_{rand} \gets  rand() $}
		\If {$X_{nearest}\gets morton\_nn(X_{rand}) == \emptyset $}
		\State {$X_{nearest}\gets nearest(X_{rand}) $}
		\EndIf
		\State {$X_{new}\gets steer(X_{nearest}, X_{rand}) $}
		\State {$state\gets morton\_collision(X_{new}, X_{nearest})$}
		\If {$state !=NO\_COLLISION$}
		\State {$state\gets collision(X_{new})$}
		\EndIf
		\State {$morton\_update(state, X_{new})$}
		\If {$collision(X_{new}, X_{nearest})==NO\_COLLISION $ }
		\State {$nn.add(X_{new})$}
		\EndIf
		\EndWhile
	\end{algorithmic}
\end{algorithm}



\noindent{\bf Hardware-Assisted Memoization. } The software-based Morton store reduces the 
execution time significantly by skipping 
 time-consuming baseline NN search and collision detection, and it 
 can be further accelerated by hardware-assisted memoization. Since Morton code calculation and 
 matching is processed by 
 iterating all the existing Morton codes sequentially in software, it can take multiple instructions. 
  To estimate how 
 many dynamic instructions
 the Morton store takes, we use Valgrind \cite{valgrind} to count the 
 number of dynamic intructions of 
 the Morton store lookups and updates. 
 \begin{figure}[h]
 	\centering
 	\includegraphics[width=0.99\linewidth]{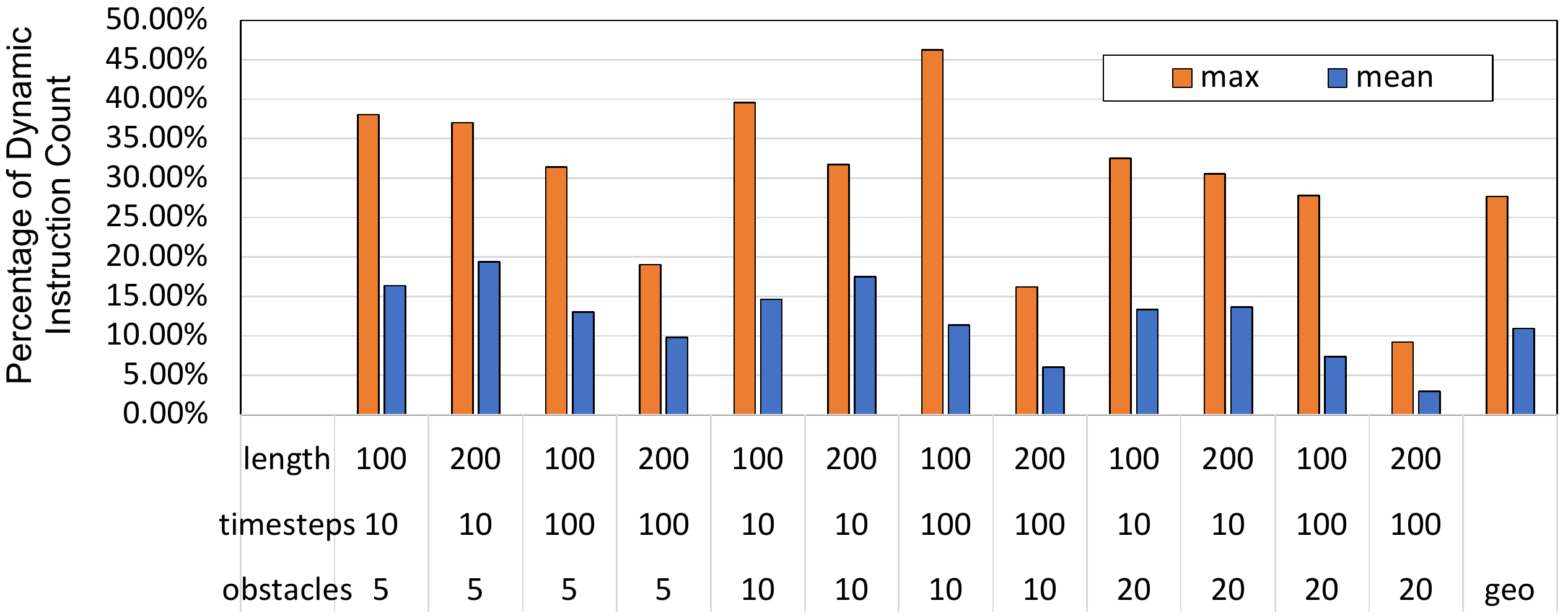}
 	\caption{Percentage of dynamic instruction count of the Morton store lookups and 
 	updates, geomean is shown in the last column. }\label{fig:morton}
 	\vspace{-0.25cm}
 \end{figure}
 
 Figure~\ref{fig:morton} shows the profiled results of different testcases with different map edge 
 length,  timesteps, and number of 
 dynamic obstacles. The geomean shows that
 Morton store-related dynamic instruction count accounts for ~27\% of the maximum total dynamic 
 instructions, and in the 
 worst case, it 
 can be 
 more than 45\% of the total number of dynamic instructions. 
  However, with hardware-assisted memoization, the dynamic instruction counts can be reduced 
  dramatically. 
We can use content-addressable memory for the Morton store lookup and update
 directly in hardware to reduce the dynamic instruction count and the correspondingly the execution 
 time.
 
 \begin{figure}
 	\vspace{-0.25cm}
	\centering
	\includegraphics[width=0.99\linewidth]{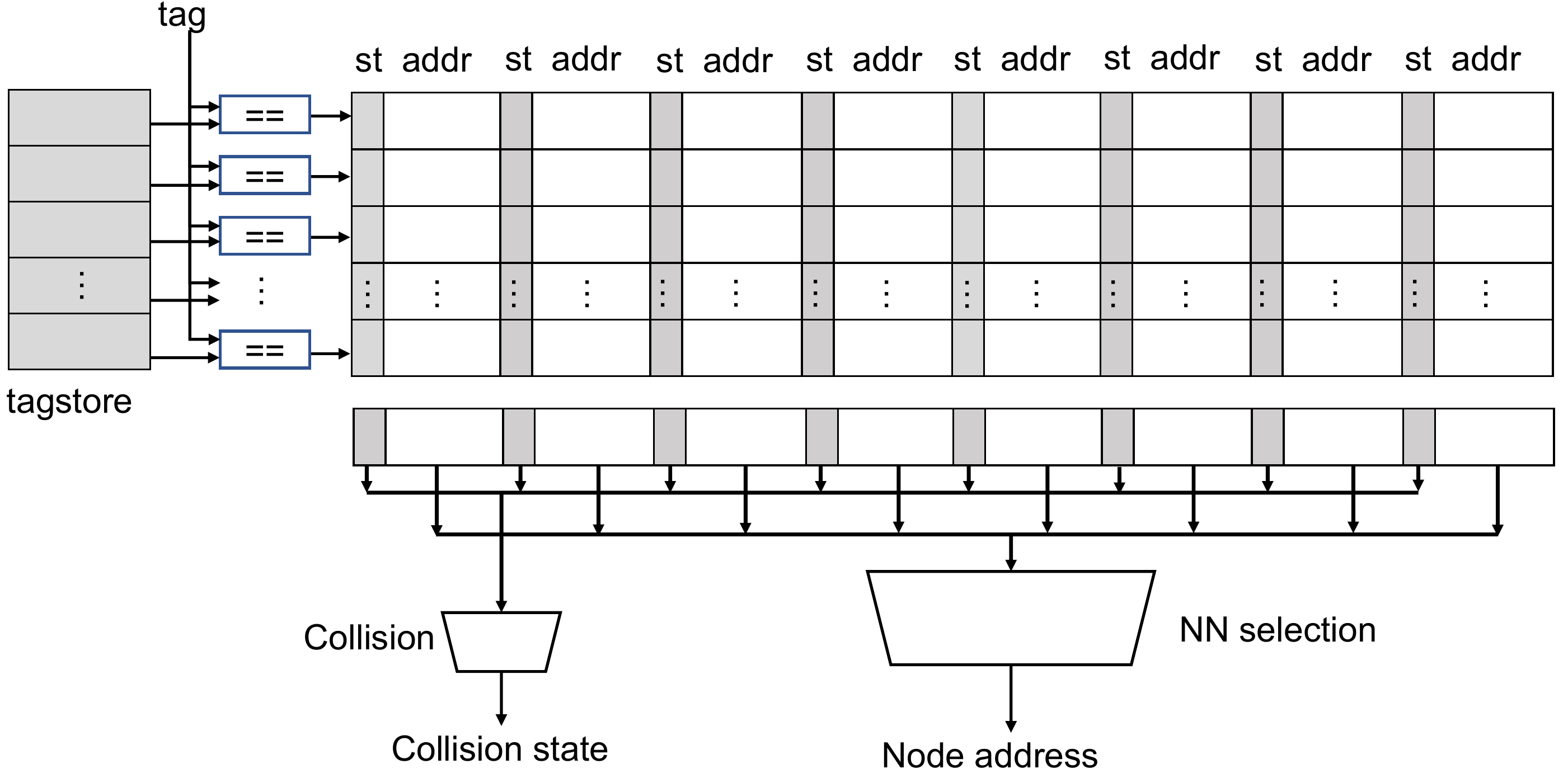}
	\vspace{-0.75cm}
	\caption{HW-based Morton store implemented by content-addressable 
	memory.}\label{fig:mem}
		\vspace{-0.25cm}
\end{figure}

\begin{figure}[h]
	\centering
	\includegraphics[width=0.6\linewidth]{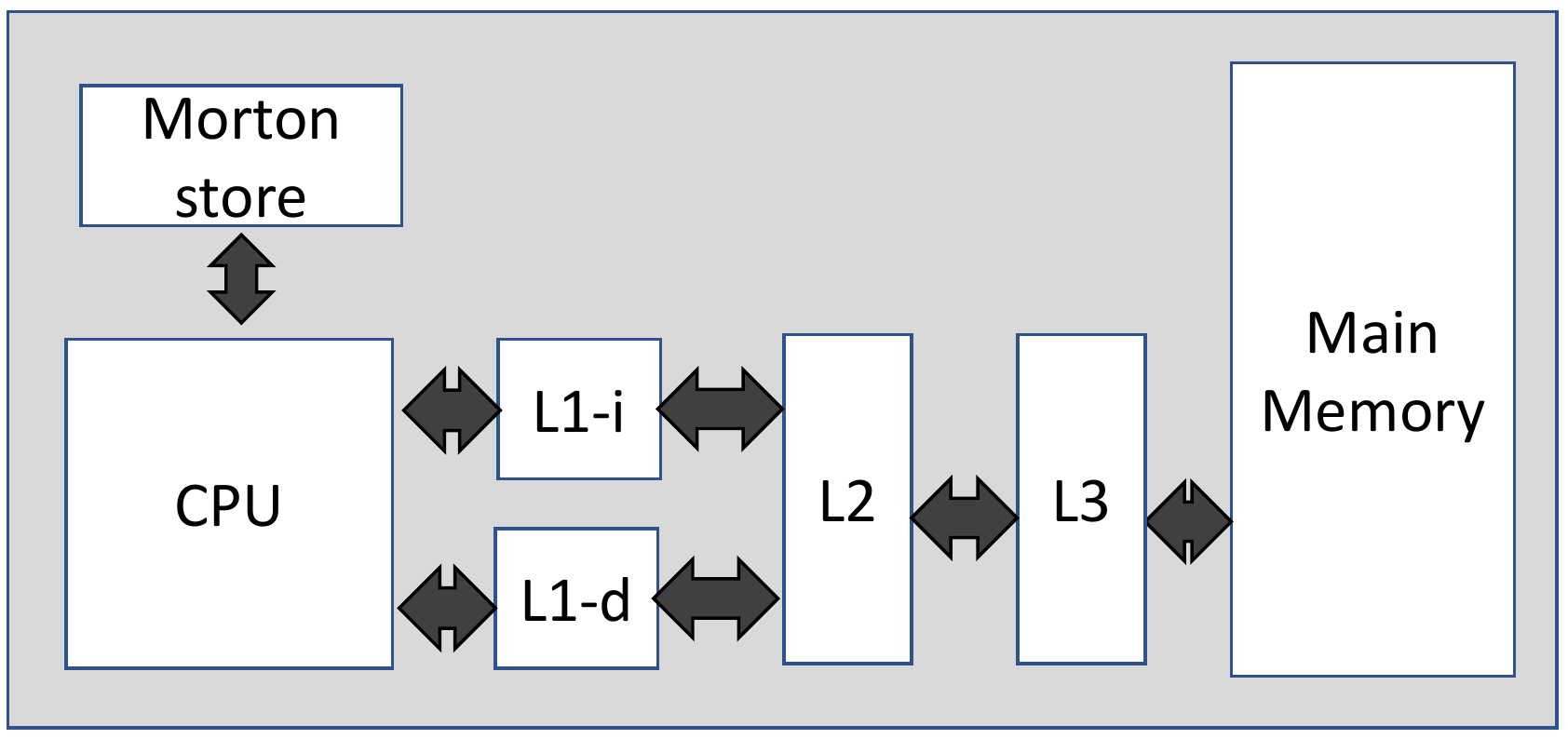}
	\vspace{-0.25cm}
	\caption{Overall architecture with Morton store.}\label{fig:arch}
	\vspace{-0.5cm}
\end{figure}

 The hardware-based Morton store  is implemented by a fully-associative content-addressable 
 memory 
 whose memoryline size is 64 bytes. Each line can store eight 8-byte addresses of RRT nodes. 
 Since these nodes are allocated on the stack, the MSBs are always 0 in our settings, and 
 we use the most significant 8 bits of the address to indicate whether there is a collision or not. Figure~\ref{fig:mem} 
 shows an example of the hardware Morton store implemented in content-addressable memory
 The tagstore keeps the 
 Morton codes, and each row contains the state (collision/no collision) information as well as the 
 addresses of the nodes. For collision state, as long as there is one state that indicates collision among all states in 
 one memoryline, the output will be a collision. 

 The content-addressable memory is connected directly to the CPU, as shown in
Figure~\ref{fig:arch}.  On a hit, the 
 corresponding memoryline used for extracting collision state or a node address is processed. 
 On a read miss, nothing is modified in the Morton store. On a write miss, the oldest 
 referenced memoryline is evicted.
 In order to access this content-addressable memory in software and provide flexibilty to different planning 
 algorithms, we define the following ISA 
 extensions: \texttt{morton\_update}, \texttt{morton\_col} and \texttt{morton\_nn} which are listed in 
 Table~\ref{tab:isa}.
 \texttt{morton\_update} takes in the coordinates of a node and its collision state and updates this 
information in the Morton store. \texttt{morton\_col} takes in the coordinates of a 
node, look it up in the content-addressable memory and decides whether there is a collision. 
\texttt{morton\_nn} finds the memory address of the approximate nearest 
neighbor in the Morton store by looking up entry with the same Morton code.
In this work, we use these instructions to accelerate RRT. However, it can also be used in other 
sampling-based planning 
algorithms such as PRM.

 \begin{table*}
 	\footnotesize
 	\centering
 	\caption{ISA interface for Morton store.}\label{tab:isa}
 	 	\vspace{-0.25cm}
 	\begin{tabular}{|l|l|l|}
 		\hline
 		\bf ISA Instruction & \bf Operation\\ \hline\hline
 		\texttt{morton\_update <x|y>, <t>, <addr>} &     Update a node with coordinate $(x,y, t)$ and 
 		memory address \texttt{addr} in Morton store \\ \hline
 		\texttt{morton\_col <x|y>, <t>, <st>}&    Check whether there is a collision for the node $(x,y,t)$ 
 		using 
 		Morton 
 		store, returns result in \texttt{st} \\ \hline
 		\texttt{morton\_nn <x|y>, <t>, <addr> }& Find an approximate near neighbor of the node  $(x,y, 
 		t)$, returns the 
 		address in \texttt{addr} \\ \hline
 		
 	\end{tabular}
 	\vspace{-0.25cm}
 \end{table*}

\section{Evaluation}\label{sec:evaluation}

\begin{figure}[h]
	\centering
	\includegraphics[width=0.9\linewidth]{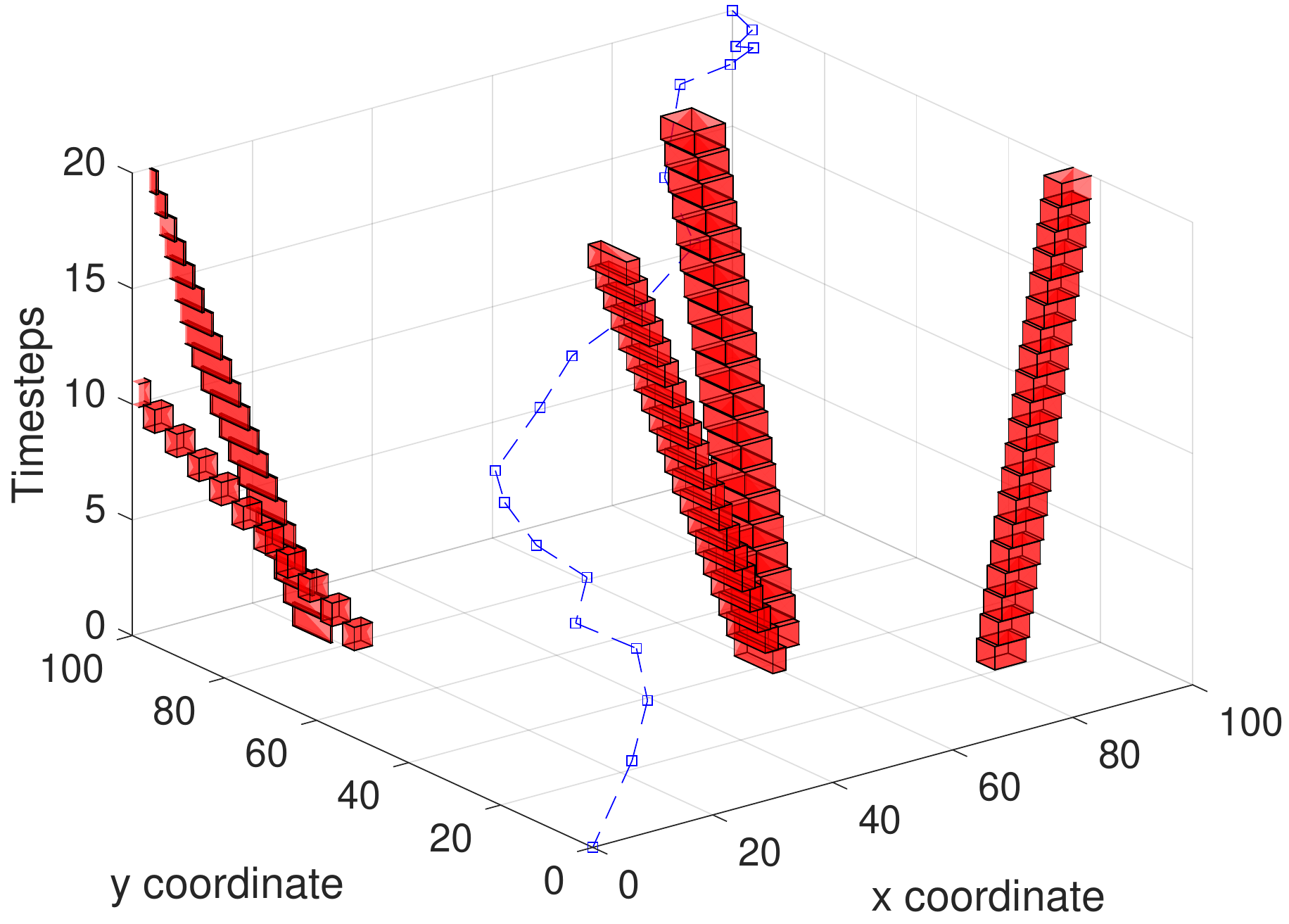}
	\caption{A test case with 5 dynamic obstacles, edge length 100, and 20 
	time steps. The solution path is shown in blue.}\label{fig:3drrt}
\vspace{-0.25cm}
\end{figure}

\noindent{\bf Synthetic Test Cases.} Each synthetic test case is defined on a square 
map, with dynamic obstacles running on it. The synthetic 
test cases are characterized 
by three 
parameters, i.e., the number of dynamic obstacles, the edge length of the square map, and 
the 
number of time steps. The solution path should start at $(0,0)$ and destination is $(l,l)$ 
where $l$ is the edge length of the square map.
We randomly generate the starting and ending locations of the obstacles at the 
beginning and the 
ending of the simulation period. The intermediate positions of the obstacles are  linearly 
interpolated. Figure~\ref{fig:3drrt} shows an example of the 
test case with 5 
dynamic obstacles, the edge length of 100, and 20 time steps. The solution path from (0,0) to 
(100,100) is 
marked in blue.

The number of obstacles determines the difficulty of the path planning problem. 
In general, with more 
obstacles, it takes more time to find a safe path and the length of the path 
may also be longer 
because more space is occupied by the obstacles.
The edge length of the map and the number of time steps mainly increase the execution 
time for path planning. On the 
other hand, 
a longer edge length and more time steps will provide more accurate solutions with 
more 
intermediate points along the 
path.

\noindent{\bf Software Performance Evaluation. }
We evaluate the  software-based Morton store  on a desktop with a 3.4GHz Intel i7-6700 
CPU and 16GB DDR4 memory at 2133MHz.
The baseline RRT algorithm is implemented in C++.  The baseline nearest neighbor 
search is based on kd-tree  \cite{kdtree}.
We use libmorton \cite{libmorton18} 
for  Morton code calculation. 

  \begin{figure}
	\centering
	\vspace{-0.5cm}
	\includegraphics[width=0.99\linewidth]{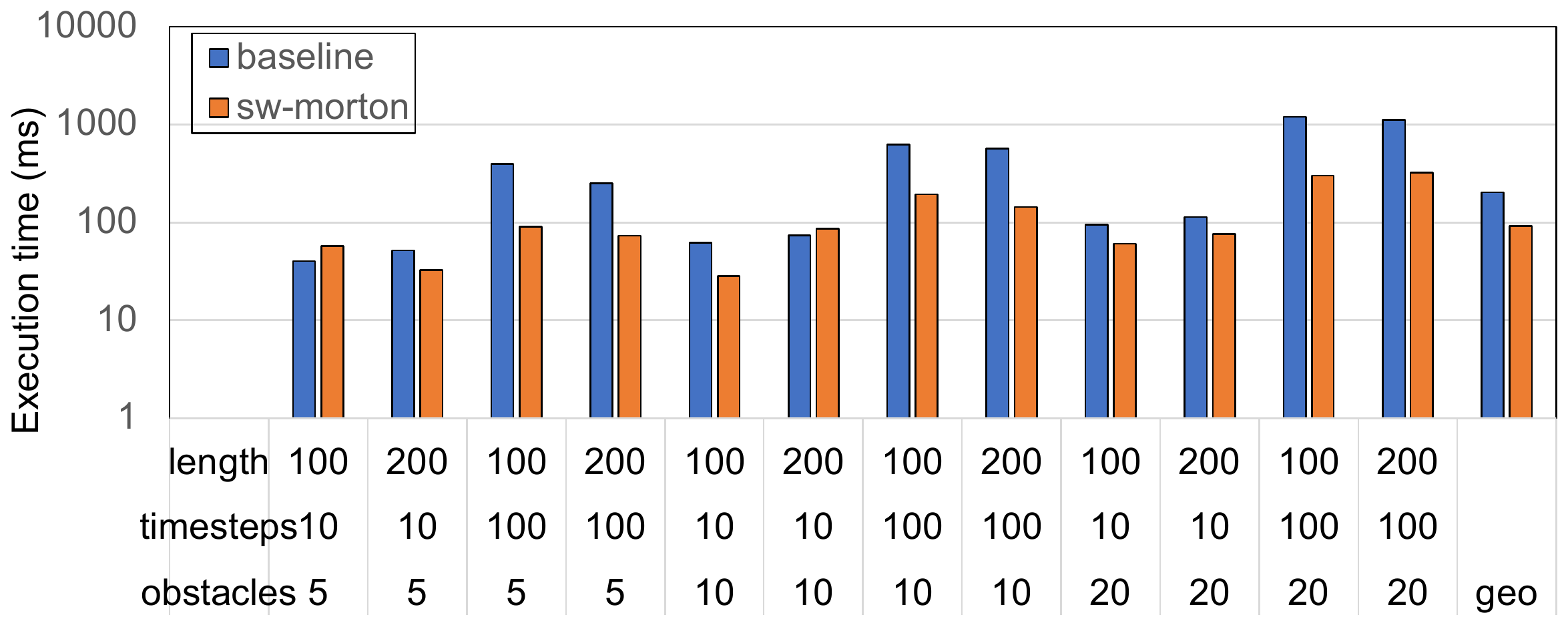}
	\caption{Execution time of the baseline and the software Morton 
		store (software on Intel i7), geomean is shown in the last column.}\label{fig:exectime}
	\vspace{-0.25cm}
\end{figure}

We simulate 12 configurations with the map edge length in \{100, 200\}, the number of time steps in 
\{10,100\}, and
the number of obstacles in \{5, 10, 20\}. 
We compare the total execution time of the baseline RRT and the RRT with the software Morton-store 
(sw-morton). We choose  $k=18$ least significant bits to mask.
For each configuration, we randomly generate 10 different test cases.  As the 
algorithm is a 
randomized 
algorithm, for each test case, we repeat each experiment 10 times and show the 
average 
execution time.
Figure~\ref{fig:exectime} shows the execution time. On average, the software Morton 
store-based path planning has 1.96$\times$  performance improvement over the baseline RRT.


\noindent{\bf Hardware Evaluation. }
The proposed hardware-assisted approach uses the content-addressable memory for 
a fast look-up and update of the recently-accessed tree nodes for the nearest neighbor search and  
collision detection.
We evaluate the performance of this hardware-based Morton store using gem5 simulator 
\cite{Binkert11thegem5}. The simulator configuration is shown in 
Table~\ref{tab:gem5}. We implemented a HW Morton store and a customized port 
between a CPU and a HW Morton store  
for the look-up and update in gem5. We choose $k=18$ least significant bits to mask.

\begin{table}
	\vspace{-0.25cm}
	\scriptsize
	\centering
	\caption{gem5 configuration.}\label{tab:gem5}
		\vspace{-0.25cm}
	\begin{tabular}{|l|l|}
		\hline
\bf Core &   Simple Timing CPU @ 2GHz  \\ \hline
\bf L1-i&   32KB, 4-way, latency=2 cycles \\ \hline
\bf L1-d&   32KB, 8-way, latency=2 cycles \\ \hline
\bf Morton Store& 32KB, fully-associative, memoryline=64 bytes, latency=2 cycles \\ \hline
\bf L2& 256KB , 8-way, latency=20 cycles \\  \hline
\bf L3&  8MB,16-way, latency=20 cycles\\ \hline
\bf Memory& 4GB, DDR3@1600MHz \\ \hline
	\end{tabular}
\vspace{-0.25cm}
\end{table}

\begin{figure}
			\centering
	\includegraphics[width=0.45\columnwidth]{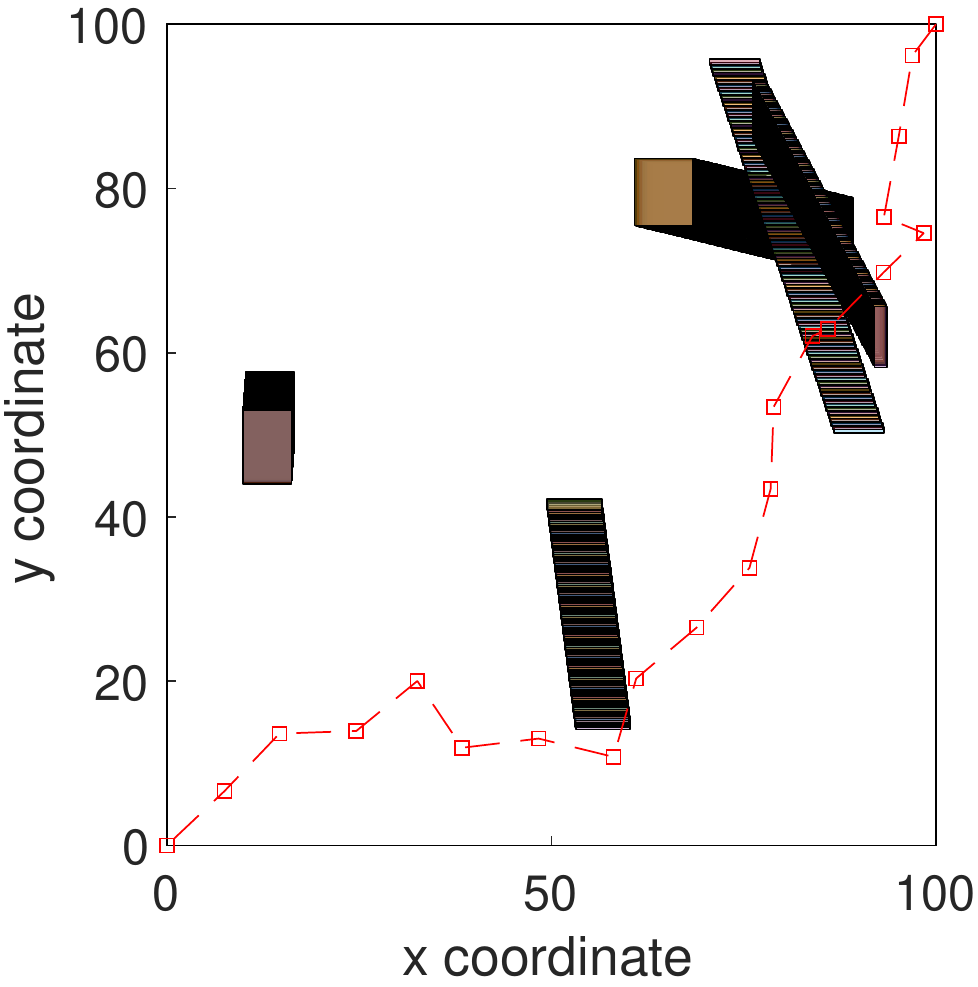}
	\caption{Solution path on the map by baseline RRT with length 175.}
	\label{fig:baseline}
	\vspace{-0.25cm}
\end{figure}

\begin{figure}
		\centering
\includegraphics[width=0.45\columnwidth]{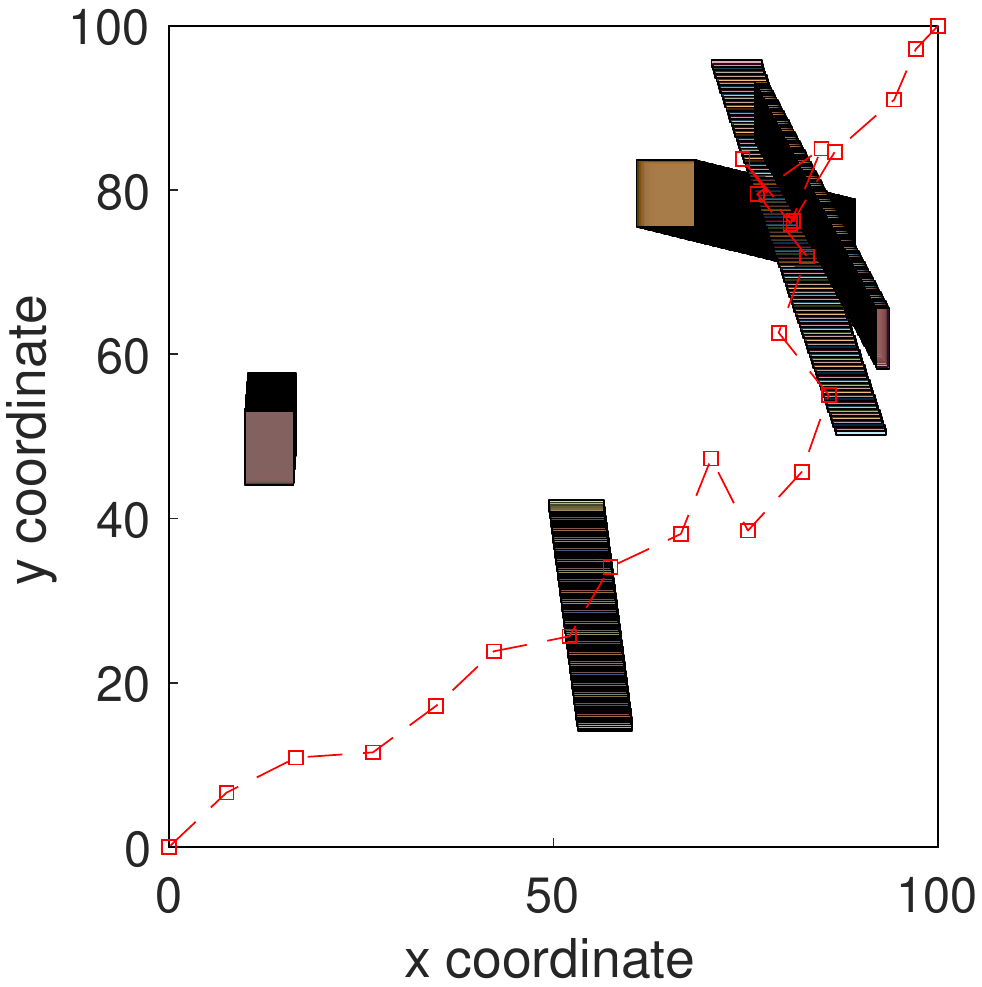}
\caption{Solution path by the HW-based Morton store with length 217.}
\label{fig:hw-morton}
\vspace{-0.25cm}
\end{figure}

\begin{figure}
		\centering
\includegraphics[width=0.99\linewidth]{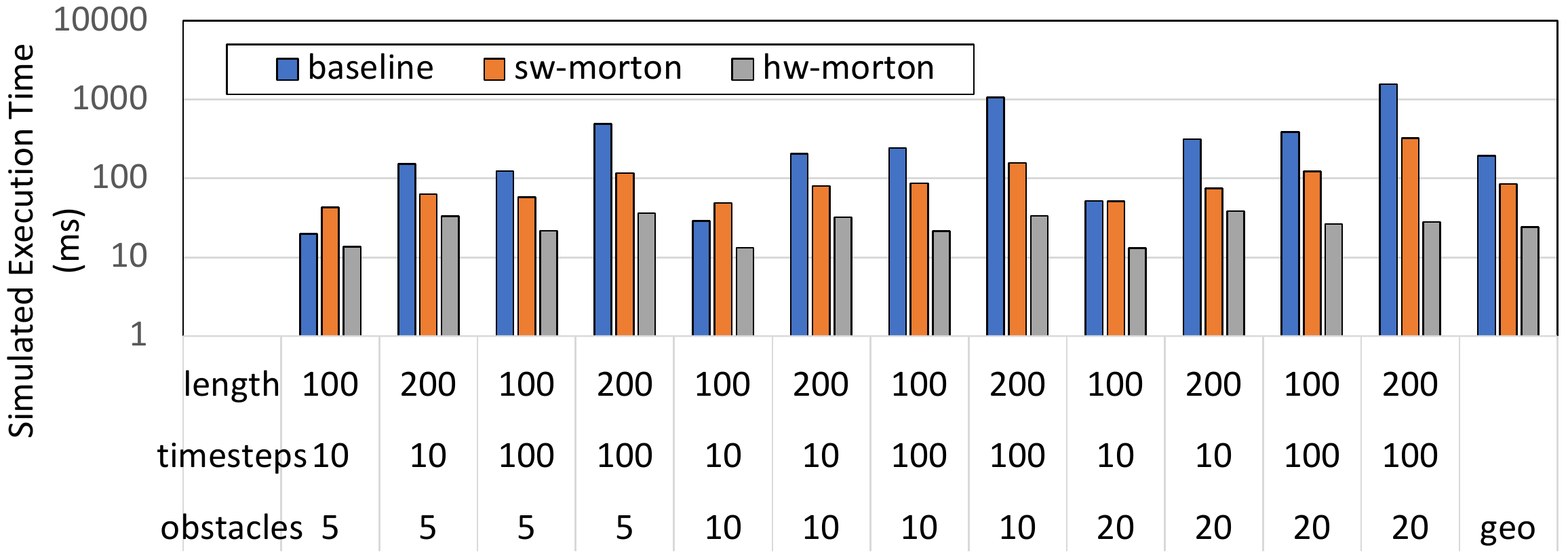}
\vspace{-0.5cm}
\caption{Execution time comparison (gem5 simulation results), geomean is shown in the last column. 
}\label{fig:time}
\vspace{-0.5cm}
\end{figure}

\begin{figure}
	\centering
	\includegraphics[width=0.99\linewidth]{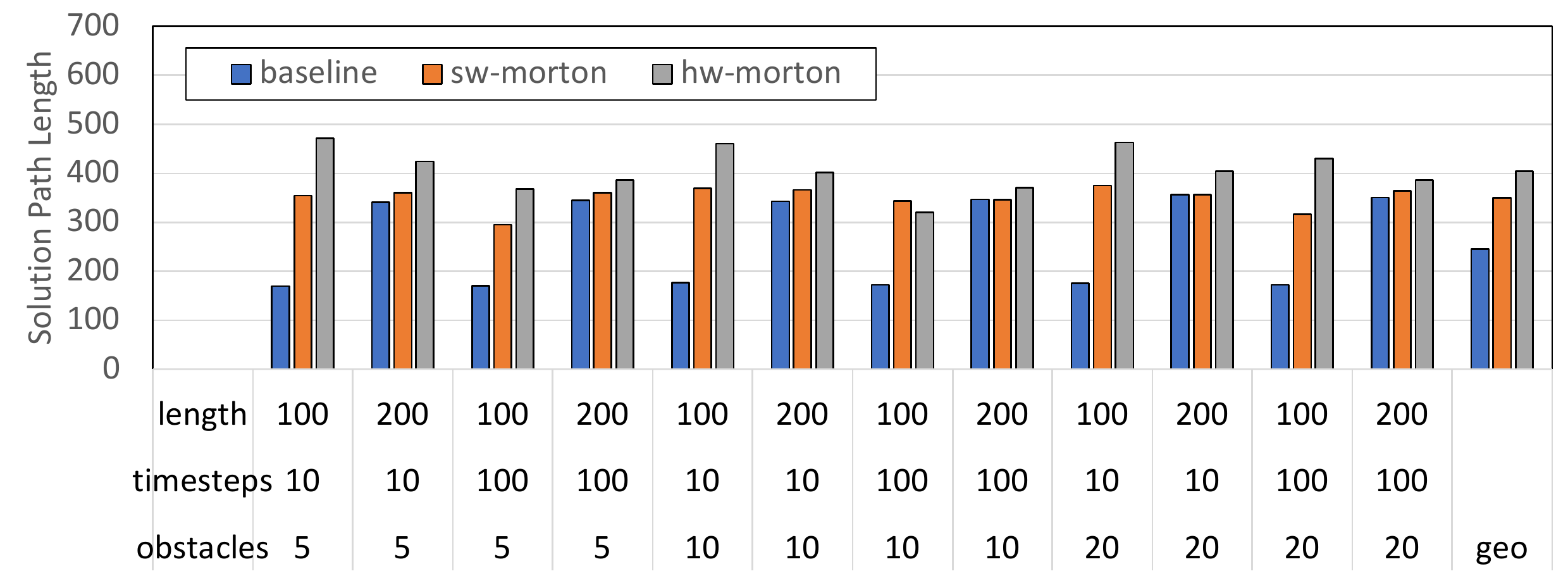}
	\vspace{-0.5cm}
	\caption{Solution length comparison, geomean is shown in the last column. }\label{fig:length}
	\vspace{-0.5cm}
\end{figure}

We use the same 12 configurations that are used in the software evaluation in the 
previous subsection.
Figure~\ref{fig:time} shows the execution time of the baseline, the software-based Morton  
store (sw-morton)
 and the hardware-based Morton (hw-morton) store based on gem5 simulation.
The performance improvements over the baseline are 2.28$\times$  for software-based Morton store 
(which is 
similar to 1.96$\times$  in software simulation using Intel i7 processor.) and 8$\times$  for 
the hardware-based Morton store on average. 
For the best 
 configuration, the performance is improved by 6.74$\times$  and 56.3$\times$  for software 
 and 
 hardware Morton stores,
 respectively.
Figure~\ref{fig:length} shows the solution path lengths for the baseline, the software-based 
Morton store and 
the hardware-based Morton store. The average solution length of the software Morton 
and hardware 
Morton stores  
are 1.42$\times$  and 1.65$\times$  higher than that of the baseline solution. This is due to the fact 
that 
the SW or HW Morton store can only find approximate nearest neighbor instead 
of the exact nearest neighbor. Thus, the solution path may fluctuate more compared to the baseline.
As shown in Figure~\ref{fig:hw-morton}, around position $(80, 80)$ of the map, the path 
given by the HW-based Morton store moves back and forth before 
reaching $(100, 100)$, which increases the solution length compared to the baseline 
solution shown in Figure~\ref{fig:baseline}. Such path can use refinement 
\cite{alterovitz2011rapidly} 
to reduce the overall length.

\begin{figure}
	\centering
	\begin{minipage}[t]{0.41\linewidth} 
		\centering
		\includegraphics[width=0.65\linewidth]{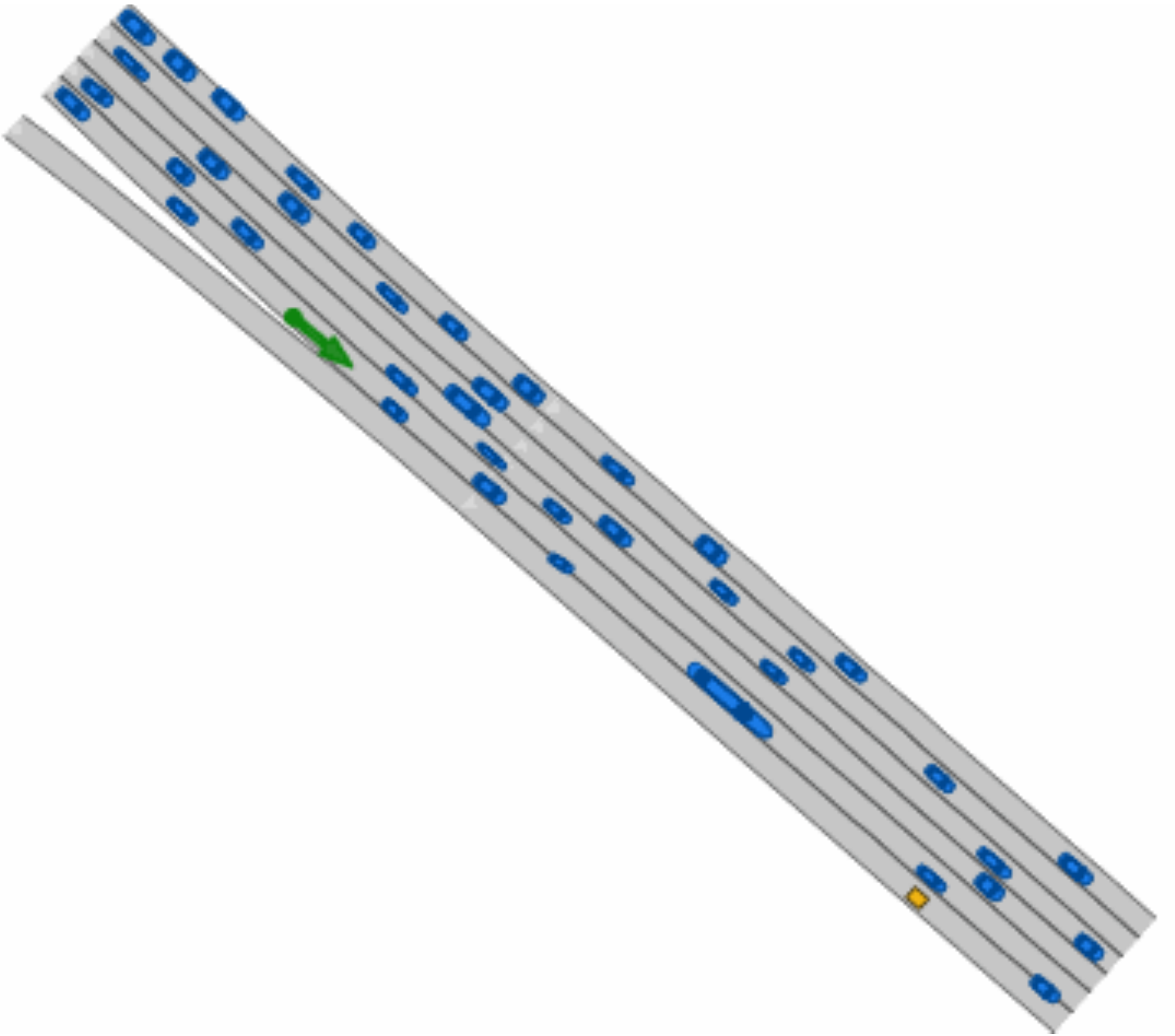}
		\vspace{-0.25cm}
		\caption{\texttt{USA\_US101-20\_1\_T\_1} from Commonroad 
			benchmarks. Blue blocks are moving obstacles, grey stripes are the lanes, green arrow is the 
			initial 
			position and yellow dot is the destination.}\label{fig:usa_us101}
		\vspace{-0.25cm}
	\end{minipage}  
	\hspace{0.25cm}
	\begin{minipage}[t]{0.49\linewidth}
		\centering
		\includegraphics[width=\textwidth]{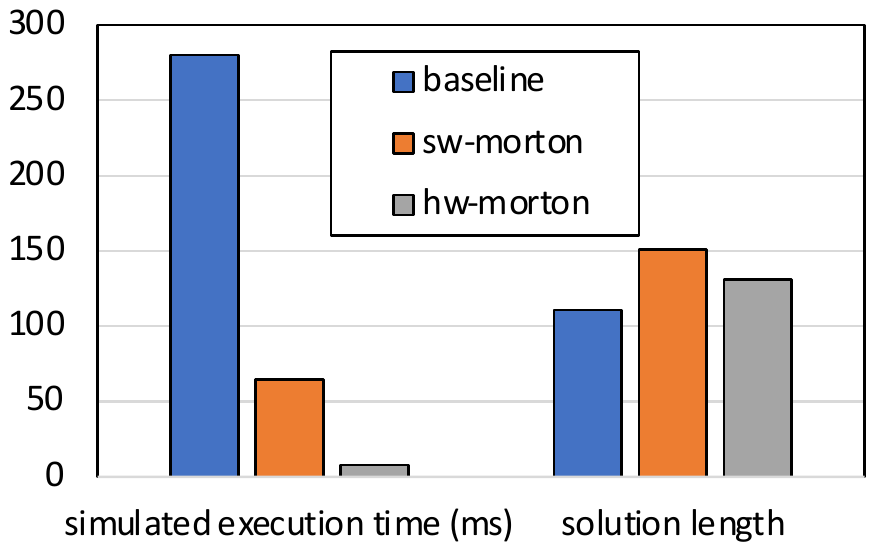}
		\vspace{-0.25cm}
		\caption{Execution time and solution length comparison of   
			\texttt{USA\_US101-20\_1\_T\_1}, sw- and hw-morton achieves 4.3$\times$  and 35$\times$  
			execution time 
			reductions.}
		\label{fig:usa_us101_perf}
		\vspace{-0.25cm}
	\end{minipage}
	\vspace{-0.25cm}
\end{figure}

\noindent{\bf Commonroad Evaluation.} To demonstrate the effectiveness of the proposed method on 
realistic scenarios, we use one example (\texttt{USA\_US101-20\_1\_T\_1}) shown in 
Figure~\ref{fig:usa_us101} containing dynamic 
obstacles from the Commonroad 
\cite{althoff2017commonroad} benchmarks, whose scenarios are partly recorded from real traffic. The 
execution time and solution path lengths are shown in Figure~\ref{fig:usa_us101_perf}, which confirms 
effectiveness in reducing the execution time.


\begin{table*}[h]
	\centering
	\scriptsize
	\caption{Comparison with existing works.}\label{tab:compare}
	\vspace{-0.25cm}
	\begin{tabular}{|l|| l|l|l|l|l|l| l|}
		\hline
		work &  implementation & indexing  &  reported improvement  & 
		algorithm & dynamic &accelerate& accelerate\\ 
		& & method & over SW& flexibility& obstacles &tree search& collision detection\\ 
		\hline\hline
		\cite{pan2016fast} &  software & LSH  & 1.46$\times$    &  yes& no & no& yes \\ \hline
		\cite{yang2020accelerating} & accelerator & RENE \cite{bremler2018encoding}  
		&  
		235$\times$ &  no& no &yes&yes \\ \hline
		\cite{kim2017brain} & processor & Zobrist \cite{zobrist1990new} & 8.63$\times$  & 
		no & 
		no&yes& no\\ \hline
		this work &  ISA extension &   Morton & 8$\times$  &  yes &yes&yes&yes\\ \hline
	\end{tabular}
	\label{tab:comparion}
	\vspace{-0.25cm}
\end{table*}

\section{Related Work}\label{sec:related_work}

There are multiple
previous studies on hardware planning accelerators. 
For example, Murray et al.\cite{murray2016microarchitecture} 
propose an accelerator for motion planning for robot arms. 
Using an ASIC, Lian et al.\cite{lian2018dadu} create a motion planning accelerator based on octree 
where 
the 
map can be updated according to the environment. 
A hybrid RRT architecture is proposed in 
\cite{malik2016fpga} and implemented in FPGA. In \cite{xiao2017parallel} 
an FPGA accelerator for RRT* is implemented.  
Unlike an accelerator or a customized processor, our approach is based on 
hardware-based memoization as ISA extension.

The content-addressable memory for path planning is also used in 
\cite{yang2020accelerating}. However, our work is different from it in several ways. 
First, our work has just one 
unified Morton store that is used  for both collision detection and nearest neighbor 
search, while 
the previous work uses 
two separate stores. Second, our work is designed as an ISA extension to a 
normal processor, 
and adapt to 
different algorithms, while the previous work \cite{yang2020accelerating} is limited
to a fixed algorithm with the size of 
the 
problem that can 
fit into the ternary-CAM available on chip. 
Kim et. al. \cite{kim2017brain} also uses CAM for path planning algorithms. 
However, it only targets the A* search algorithm, not sampling-based path 
planning. 
Also, the previous work cannot completely 
skip the tree search and does not explicitly accelerate collision detection.
Table~\ref{tab:compare} summarizes 
the major differences between our work and the related work on pruning and 
CAM-based acceleration.





\section{Conclusion}\label{sec:conclusion}


In this paper, we proposed a software-hardware co-design approach to accelerate
path planning for autonomous driving. At the algorithm/software level, we use Morton codes as the 
tags to index the approximate nearest 
neighbor and collision detection. In hardware, we use content-addressable 
memory to efficiently search for nodes indexed by Morton codes. The experimental results show that
the proposed acceleration can significantly improve the performance over the baseline, by 8$\times$ 
on average.



\section*{Acknowledgment}
The authors would like to thank Helena Caminal, Dr. Jacopo Banfi and Dr. 
Mohamed 
Ismail for the helpful discussions. This project is partially funded by NSF grant 
ECCS-1932501.


\bibliographystyle{abbrv}
\scriptsize
\bibliography{main}

\begin{thebibliography}{10}

\bibitem{alterovitz2011rapidly}
R.~Alterovitz, S.~Patil, and A.~Derbakova.
\newblock Rapidly-exploring roadmaps: Weighing exploration vs. refinement in
  optimal motion planning.
\newblock In {\em 2011 IEEE International Conference on Robotics and
  Automation}, pages 3706--3712. IEEE, 2011.

\bibitem{althoff2017commonroad}
M.~Althoff, M.~Koschi, and S.~Manzinger.
\newblock Commonroad: Composable benchmarks for motion planning on roads.
\newblock In {\em 2017 IEEE Intelligent Vehicles Symposium (IV)}, pages
  719--726. IEEE, 2017.

\bibitem{libmorton18}
J.~Baert.
\newblock https://github.com/forceflow/libmorton, 2018.

\bibitem{bremler2018encoding}
A.~Bremler-Barr, Y.~Harchol, D.~Hay, and Y.~Hel-Or.
\newblock Encoding short ranges in tcam without expansion: Efficient algorithm
  and applications.
\newblock {\em IEEE/ACM Transactions On Networking}, 26(2):835--850, 2018.

\bibitem{open2017discrete}
O.~G. Consortium et~al.
\newblock Discrete global grid systems abstract specification, 2017.

\bibitem{di2015energy}
C.~Di~Franco and G.~Buttazzo.
\newblock Energy-aware coverage path planning of uavs.
\newblock In {\em 2015 IEEE international conference on autonomous robot
  systems and competitions}, pages 111--117. IEEE, 2015.

\bibitem{Binkert11thegem5}
N.~B. et~al.
\newblock The gem5 simulator.
\newblock {\em SIGARCH Comput. Archit. News}, 2011.

\bibitem{hardy2013contingency}
J.~Hardy and M.~Campbell.
\newblock Contingency planning over probabilistic obstacle predictions for
  autonomous road vehicles.
\newblock {\em IEEE Transactions on Robotics}, 29(4):913--929, 2013.

\bibitem{ichter2017robust}
B.~Ichter, B.~Landry, E.~Schmerling, and M.~Pavone.
\newblock Robust motion planning via perception-aware multiobjective search on
  gpus.
\newblock 2017.

\bibitem{kim2017brain}
Y.~Kim, D.~Shin, J.~Lee, and H.-J. Yoo.
\newblock Brain: A low-power deep search engine for autonomous robots.
\newblock {\em IEEE Micro}, 37(5):11--19, 2017.

\bibitem{lian2018dadu}
S.~Lian, Y.~Han, X.~Chen, Y.~Wang, and H.~Xiao.
\newblock Dadu-p: A scalable accelerator for robot motion planning in a dynamic
  environment.
\newblock In {\em DAC}, pages 1--6. IEEE, 2018.

\bibitem{lin2018architectural}
S.-C. Lin, Y.~Zhang, C.-H. Hsu, M.~Skach, M.~E. Haque, L.~Tang, and J.~Mars.
\newblock The architectural implications of autonomous driving: Constraints and
  acceleration.
\newblock In {\em ASPLOS}, pages 751--766, 2018.

\bibitem{malik2016fpga}
G.~Malik, K.~Gupta, R.~Dharani, and K.~M. Krishna.
\newblock Fpga based hybrid architecture for parallelizing rrt.
\newblock {\em arXiv preprint arXiv:1607.05704}, 2016.

\bibitem{murray2019programmable}
S.~Murray, W.~Floyd-Jones, G.~Konidaris, and D.~J. Sorin.
\newblock A programmable architecture for robot motion planning acceleration.
\newblock In {\em ASAP}, volume 2160, pages 185--188. IEEE, 2019.

\bibitem{murray2016microarchitecture}
S.~Murray, W.~Floyd-Jones, Y.~Qi, G.~Konidaris, and D.~J. Sorin.
\newblock The microarchitecture of a real-time robot motion planning
  accelerator.
\newblock In {\em MICRO}, pages 1--12. IEEE, 2016.

\bibitem{pan2016fast}
J.~Pan and D.~Manocha.
\newblock Fast probabilistic collision checking for sampling-based motion
  planning using locality-sensitive hashing.
\newblock {\em The International Journal of Robotics Research},
  35(12):1477--1496, 2016.

\bibitem{kdtree}
J.~Tsiombikas.
\newblock http://nuclear.mutantstargoat.com/sw/kdtree.

\bibitem{valgrind}
Valgrind.
\newblock https://valgrind.org.

\bibitem{xiao2017parallel}
S.~Xiao, N.~Bergmann, and A.~Postula.
\newblock Parallel rrt-star architecture design for motion planning.
\newblock In {\em FPL}, pages 1--4. IEEE, 2017.

\bibitem{yang2020accelerating}
Y.~Yang, S.~Lian, X.~Chen, and Y.~Han.
\newblock Accelerating rrt motion planning using tcam.
\newblock In {\em GLSVLSI}, pages 481--486, 2020.

\bibitem{yu2020building}
B.~Yu, W.~Hu, L.~Xu, J.~Tang, S.~Liu, and Y.~Zhu.
\newblock Building the computing system for autonomous micromobility vehicles:
  Design constraints and architectural optimizations.
\newblock In {\em MICRO}, 2020.

\bibitem{zobrist1990new}
A.~L. Zobrist.
\newblock A new hashing method with application for game playing.
\newblock {\em ICGA Journal}, 13(2):69--73, 1990.

\end{thebibliography}

\end{document}